\documentclass[conference]{IEEEtran}
\IEEEoverridecommandlockouts
\usepackage{xurl}
\usepackage[utf8]{inputenc}
\usepackage[T5]{fontenc}
\usepackage{hyperref}
\usepackage{multirow}

\usepackage{cite}
\usepackage{amsmath,amssymb,amsfonts}
\usepackage{algorithmic}
\usepackage{graphicx}
\usepackage{textcomp}
\usepackage{xcolor}
\usepackage{float}
\usepackage{authblk}

\def\BibTeX{{\rm B\kern-.05em{\sc i\kern-.025em b}\kern-.08em
    T\kern-.1667em\lower.7ex\hbox{E}\kern-.125emX}}
\begin{document}

\title{Improving Sentiment Analysis By Emotion Lexicon Approach on Vietnamese Texts}

\author[1,2,*]{An Long Doan}
\author[1,2,†]{Son T. Luu}
\affil[1]{University of Information Technology, Ho Chi Minh City, Vietnam}
\affil[2]{Vietnam National University, Ho Chi Minh City, Vietnam}

\affil[ ]{Email: *\textit {19521173@gm.uit.edu.vn}, †\textit{sonlt@uit.edu.vn}}

\maketitle

\begin{abstract}
The sentiment analysis task has various applications in practice. In the sentiment analysis task, words and phrases that represent positive and negative emotions are important. Finding out the words that represent the emotion from the text can improve the performance of the classification models for the sentiment analysis task. In this paper, we propose a methodology that combines the emotion lexicon with the classification model to enhance the accuracy of the models. Our experimental results show that the emotion lexicon combined with the classification model improves the performance of models. 
\end{abstract}


\section{Introduction}
\label{intro}
The topic of sentiment analysis (SA) has attracted a lot of academic interest and research, particularly in the development of predictive models. SA has various applications in daily life since it is a tool to monitor opinions from user-generated data and assist decision-making \cite{BIRJALI2021107134}. The application of sentiment analysis appeared in many fields, such as e-commerce, social media, blogs, discussion forums, and education. 

In the SA tasks, words and phrases which represent the negative and positive sentiments play essential roles \cite{liu2012sentiment}. According to \cite{10.1145/3057270}, the lexicon methods try to find the "prior polarity" meaning of the word, while the machine learning methods try to create a generic classifier from the domain specified labeled dataset with the purpose of extracting the "contextual priority" from the text. Those methodologies have advantages and disadvantages since the authors in \cite{10.1145/3057270} propose an approach to combine both two methodologies to improve the performance of the sentiment classifier. Therefore, in this paper, we propose a methodology of integrating the emotion lexicons with the machine learning models to enhance the performance of the classifiers for the sentiment analysis task in the Vietnamese language. 

Previous works in Vietnamese sentiment tasks have created the dataset on specific domains such as social networks, education, e-commerce, and the emotion lexicon. In this paper, we use three datasets, including UIT-VSMEC \cite{vsmec}, UIT-VSFC \cite{vsfc}, and ViHSD \cite{vihsd} with the VnEmoLex \cite{VnEmolex2017} to investigate the performance of our approach on the Vietnamese sentiment analysis task. All three datasets are the large-scale dataset and are manually annotated by humans with a strict annotation procedure on a specific domain such as the social media domain (UIT-VSMEC), student and education domain (UIT-VSFC), and hate speech detection (ViHSD). Besides, VnEmoLex is a lexicon emotion set with eight different emotion types and contains 12,795 emotional words.  

Our paper is structured as follows. Section \ref{related} surveys several current works in the Vietnamese sentiment analysis task. Section \ref{dataset} takes a brief look at the used datasets for our experiments. Section \ref{method} describes our proposed methodologies to combine the lexicon features with the machine learning classifier. Section \ref{ex} illustrates our experimental results and describes the error analysis for the proposed method. Finally, section \ref{conclusion} concludes our works and suggests future works. 

\section{Related Works}
\label{related}

The sentiment analysis task can be categorized as the text classification task. Various benchmark datasets are created to serve the sentiment analysis task in Vietnamese for different domains, such as the VLSP 2018\cite{nguyen2018vlsp} and UIT-ABSA\cite{van2021two} dataset for aspect-based Sentiment Analysis on restaurant and hotel domains, the UIT-VSFC\cite{vsfc} dataset for sentiment analysis on student feedback, the UIT-VSMEC\cite{vsmec} for emotional classification of user comments on social network sites, the UIT-ViSFD\cite{10.1007/978-3-030-82147-0_53} for aspect-based sentiment analysis about smartphone feedback, and the ViHSD\cite{vihsd} and VLSP 2019 HSD\cite{vu2020hsd} dataset for hate speech detection on social media texts (According to \cite{schmidt-wiegand-2017-survey}, hate speech detection and sentiment analysis tasks are related because they both treat the negative and positive sentiment through the hate speech message). We choose the UIT-VSMEC, UIT-VSFC, and ViHSD as three datasets for evaluating our proposed methodology.

Besides the annotated dataset, VnEmoLex \cite{VnEmolex2017} and VietSentiWordNet \cite{vu2014construction} are two lexicons used for the sentiment analysis task. The VnEmoLex contains eight fundamental levels of sentiment, including joy, sadness, anger, fear, trust, disgust, surprise, and anticipation, while the VietSentiWordNet contains only three levels, which are positivity, negativity, and neutrality. In this paper, we use the VnEmoLex lexicon because it has more levels of emotion than the VietSentiWordNet. 

Finally, based on each dataset, there are several approaches to construct the classification models to detect the sentiment from text. The Maximum entropy model achieved the best result on the UIT-VSFC dataset \cite{vsfc}, the Text-CNN model obtained the highest result on the UIT-VSMEC dataset \cite{vsmec}, and the BERT model gave the best result on the ViHSD dataset \cite{vihsd}. From the current baseline models on each three datasets, we propose our methodology, which combines the emotion lexicon with the current classifier to boost the performance. 
\section{Vietnamese Sentiment Analysis Datasets}
\label{dataset}

The UIT-VSMEC is created for emotion detection on Vietnamese social media text \cite{vsmec}. This corpus has a total of seven levels of emotion as described in Table \ref{tab:1}. We use this dataset as the benchmark dataset to evaluate the efficiency of our proposed methodology. Besides the UIT-VSMEC, we also analyze our results on two remaining benchmark datasets, including UIT-VSFC and the ViHSD to empathize with the effectiveness of our methodology. The UIT-VSFC is created to analyze the feedback of students about education activity \cite{vsfc}. This corpus has two tasks: the sentiment-based task for detecting user emotion from the text about the education activity and the topic-based task for classifying the categories belonging to the teaching and learning activities such as lecturer, facility, and curriculum \cite{vsfc}. In this paper, we use the sentiment-based task for our experiments. The labels of the UIT-VSFC are shown in Table \ref{tab:1}. Finally, the ViHSD is a dataset created for the hate speech detection task on the Vietnamese language \cite{vihsd}. This corpus also has three labels as shown in Table \ref{tab:1}. All three datasets are manually annotated by humans with a detailed and strict annotation procedure. 

\begin{figure}[H]
    \centering
    \includegraphics[scale=0.6]{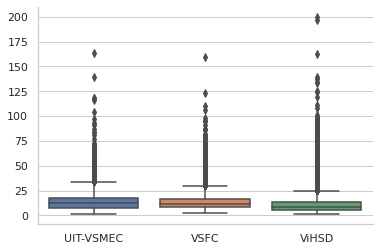}
    \caption{ Distribution of the length of the comments in the three Vietnamese Sentiment Analysis datasets.}
    \label{fig:1}
\end{figure}

Table \ref{tab:1} gives a summary of the labels for the three aforementioned datasets along with percentages and examples. The distribution of comment lengths in the UIT-VSMEC, UIT-VSFC, and ViHSD datasets is depicted in Figure \ref{fig:1}. It can be seen from Figure \ref{fig:1} that the average length of sentences of the three datasets is nearly the same, which are 14.01 for the UIT-VSMEC, 14.31 for the UIT-VSFC, and 11.51 for the ViHSD. In addition, both three datasets are imbalanced in the label distribution, according to Table \ref{tab:1}. For the UIT-VSMEC, the labels are skewed to Enjoyments, Sadness, and Disgust. For the UIT-VSFC, the labels are skewed mainly to the Positive. For the ViHSD dataset, the labels are skewed to the CLEAN label. Additionally, according to examples shown in Table \ref{tab:1}, we discovered that sentences are frequently short due to the brevity users seek to communicate in social media texts (apart from purposeful cases like spam and storytelling). Along with that, emojis and acronyms are frequently employed to speed up typing.

\begin{table}[!ht]
    \centering
    \caption{OVERVIEW STATISTICS OF THE THREE VIETNAMESE SENTIMENT ANALYSIS DATASETS.}
    \label{tab:1}
    \begin{tabular}{lcclc}
    \hline
    \multicolumn{1}{c}{\textbf{Dataset}} & \multicolumn{1}{c}{\textbf{Size}} & \multicolumn{1}{c}{\textbf{Average length}} & \multicolumn{1}{c}{\textbf{Labels}} & \multicolumn{1}{c}{\textbf{Percentage}} \\
    \hline
    \multirow{7}{*}{UIT-VSMEC} & \multirow{7}{*}{6,927} & \multirow{7}{*}{14.01} & FEAR & 5.73 \\
     &  &  & SURPRISE & 4.36  \\
     &  &  & ANGER & 7.04  \\
     &  &  & ENJOYMENT & 28.08 \\
     &  &  & SADNESS & 17.07  \\
     &  &  & DISGUST & 19.32  \\
     &  &  & OTHER & 18.40  \\ 
    \hline
    \multirow{3}{*}{UIT-VSFC} & \multirow{3}{*}{16,175} & \multirow{3}{*}{14.31} & POSITIVE & 49.38 \\
     &  &  & NEGATIVE & 4.02  \\
     &  &  & NEUTRAL & 46.60  \\ 
    \hline
    \multirow{3}{*}{ViHSD} & \multirow{3}{*}{33,400} & \multirow{3}{*}{11.51} & CLEAN & 82.70  \\
     &  &  & OFFENSIVE & 6.67  \\
     &  &  & HATE & 10.63  \\ 
    \hline
    \end{tabular}
\end{table}

In general, although the three datasets have different labels because they were created for a specific domain task, they have the same feature in the text. Hence, we use these three datasets as the benchmark for evaluating the performance of our methodology. 
\section{Methodology}
\label{method}

\begin{figure}[H]
    \centering
    \includegraphics[scale=0.35]{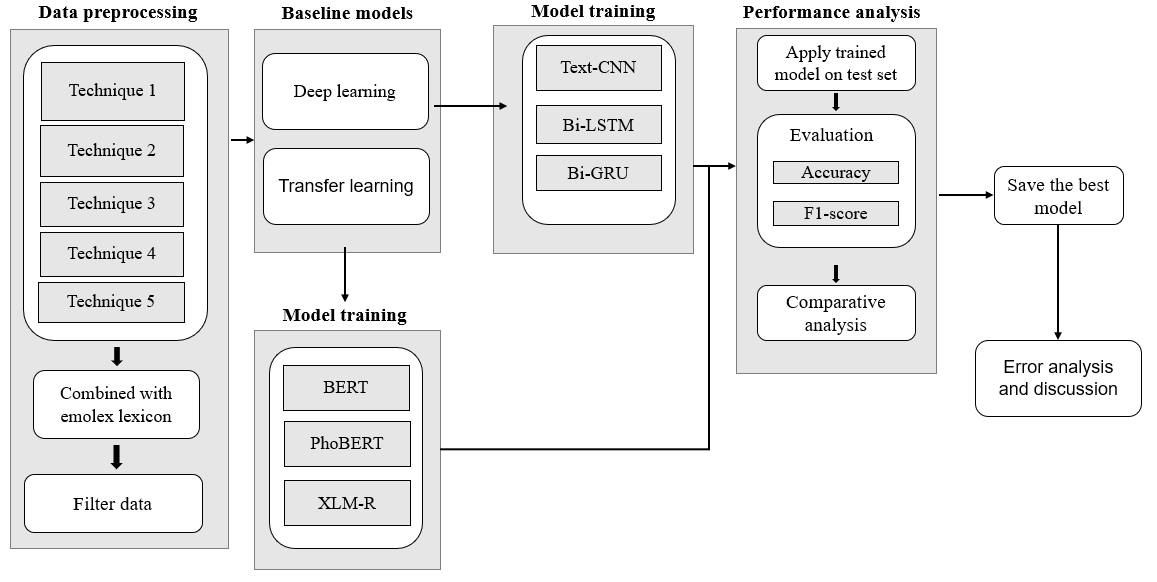}
    \caption{Experimental procedure.}
    \label{fig:3}
\end{figure}

The task of sentiment analysis is categorized as the text classification task. Figure \ref{fig:3} illustrates briefly our methodology, including pre-processing techniques, combining the emotion lexicon to the feature vectors, and fitting them to the classification models. 

\subsection{Data pre-processing}

In \cite{process}, the authors proposed seven techniques to pre-process the text based on the characteristic of Vietnamese social media texts. We adapt those pre-processing techniques for our experiments. Our pre-processing techniques are described below:  

\begin{enumerate}
    \item \textbf{Standardizing words}: Mistake words frequently appear in social media datasets, for example: “ủaaa” should be “ủa” (what?), or “đẹppp quáa” should be “đẹp quá” (so beautiful). Alternatively, you may come across numerous instances of improper punctuation, for example: “qúy hóa quá” should be “quý hóa quá” (so appreciate). In part due to the features and complexity of Vietnamese, as well as the nature of social media. As a result, in order to carry out the subsequent processing stages, the return words must be standardized. We employ the punctuation standardized, such as: “tuỳ” converted to “tùy” (depend on you), along with the word standardization technique used in \cite{process}. 

    \item \textbf{Conversion of emoticons into emoji}: For social media data sets, as well as the current state of language used by young people, capturing and processing emojis (symbols for user emotions) is necessary. Statistics from the UIT-VSMEC dataset show that more than 40\% of words contain emojis. This clearly demonstrates the social media aspect of the dataset, which makes it challenging to deal with these emoticons. Because each emoji has a unique nuance and because an emoticon is tailored to a certain emoji, for example: “:)))” or “=))”  is normalized as “:)” (slightly smiling face). The papers have also identified and developed the issue with how emojis are handled \cite{bai2019systematic,asghar2017lexicon}. Therefore, we made the decision to standardize emoticons (method 1) before switching back to emojis. In Table \ref{tab:4}, several examples of emojis conversion are shown. 

\begin{table}[H]
    \centering
    \caption{SEVERAL EXAMPLES FOR CONVERTING EMOTICONS TO EMOJIS.}
    \includegraphics[width=0.8\linewidth]{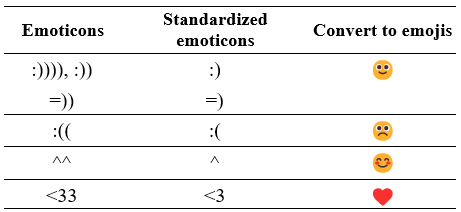}
    \label{tab:4}
\end{table}

    \item \textbf{Correction of misspelled words and search for acronyms}: Misspellings are unavoidable when people communicate on social media, especially when using isolated languages like Vietnamese. Therefore, rapid processing is always required to provide a more complete evaluation. In this paper, we use a list of Vietnamese misspellings, including 736 words that \cite{process} uses to accompany its letters. We must standardize the incorrect terms in order to replace them with the standard form in Vietnamese for each word in the list, which includes acronyms and misspelled words.

    \item \textbf{Control Teencode and Stopwords}: When performing sentiment analysis problems, Teencode \footnote{\url{https://gist.github.com/nguyenvanhieuvn/7d9441c10b3c2739499fc5a4d9ea06fb}} and Stopwords \footnote{\url{https://github.com/stopwords/vietnamese-stopwords}} are always issues that garner attention, but with Vietnamese, the level of efficacy that it provides is always at the level that must be confirmed. In order to objectively assess the effectiveness, we processed the original data set using only Teeencode and Stopwords in this paper before combining it with other techniques.

    \item \textbf{Words segmentation}: Words segmentation in Vietnamese presents a questionable issue while performing text sentiment analysis. Even though two words or phrases may share the same meaning and arouse positive emotions, segmenting them into different words causes them to have an impact and slightly dilute the meaning of the phrase and sentence. The efficiency of word segmentation for each problem, each scenario, and the target must therefore be examined using genuine intuitive experiments. We use the VnCoreNLP \cite{vu-etal-2018-vncorenlp} as the word segmenter for our experiments. 
\end{enumerate}

\subsection{Emotion Lexicons}
Annotated lexicon for the eight primary Vietnamese emotions (joy, sadness, anger, fear, trust, disgust, surprise, and anticipation) classified into two main categories, positive and negative emotions—can be found in VnEmoLex \cite{VnEmolex2017}. A total of 12,795 words make up VnEmoLex, 4,431 of which are obtained from EmoLex and the remainder 8,364 from Viet Wordnet.  

\begin{figure}[H]
    \centering
    \includegraphics[width=1\linewidth]{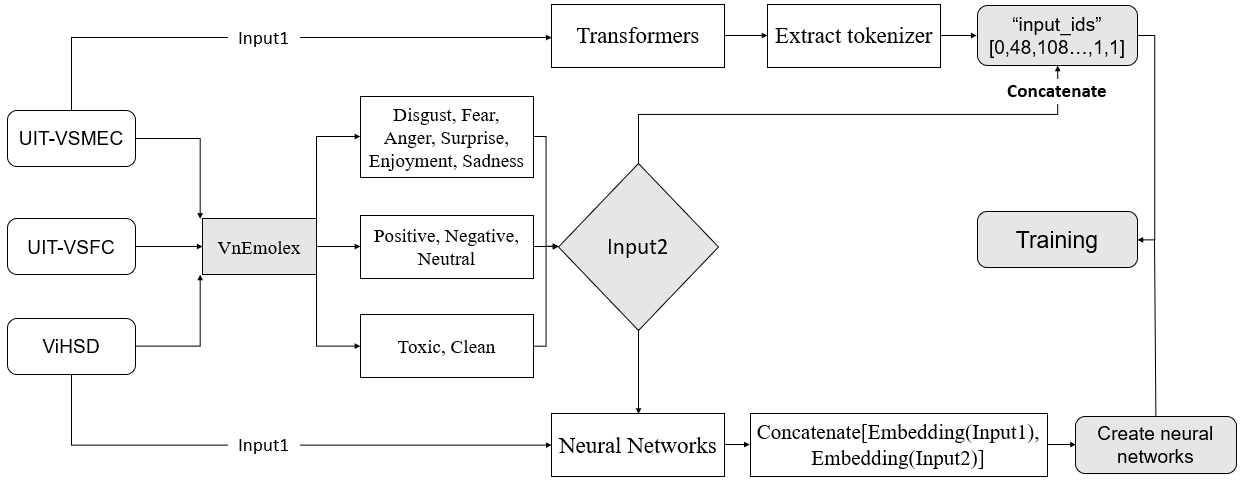}
    \caption{The procedure for putting into practice the technique of combining lexicon emotion with Vietnamese sentiment analysis.}
    \label{fig:2}
\end{figure}

The next step is to create a counter for VnEmoLex words used in sentences from the original dataset. From there, create more label attributes, which are simply the total amount of labels (emotion at VnEmolex) in the sentence. Following VnEmoLex processing, certain sentences are described in Table \ref{tab:5}. The words that appear and are represented in the emotion lexicon are marked in green. It is clear that a word can bring up a variety of feelings at once, this challenges us with a problem and makes processing words difficult. In UIT-VSMEC, six labels from VnEmoLex were chosen to fit the original dataset after being filtered out. For the UIT-VSFC, we choose the positive and negative labels in order to fit the data, other emotions outside of these two labels will be neutral. For the ViHSD dataset, we made the decision to combine the VnEmolex features and convert them into the highly adaptable in the ViHSD dataset. The labels “Anger”, “Fear” and “Disgust”, for example, will be converted to “Toxic” while the labels “Enjoyment” and “Trust” will be converted to “Clean”.


\begin{table*}[!ht]
\centering
\caption{Describe some of the sentences in the UIT-VSMEC dataset that were obtained from combining it with VnEmolex.}

\begin{tabular}{lcccccc}
\hline
\multicolumn{1}{c}{\textbf{Sentence}}                                                           & \multicolumn{1}{c}{\textbf{Disgust}} & \multicolumn{1}{c}{\textbf{Fear}} & \multicolumn{1}{c}{\textbf{Enjoyment}} & \multicolumn{1}{c}{\textbf{Sadness}} & \multicolumn{1}{c}{\textbf{Suprise}} & \multicolumn{1}{c}{\textbf{Anger}} \\ \hline
cho \colorbox{green}{đáng đời} \colorbox{green}{con quỷ} . về nhà \colorbox{green}{lôi} con nhà mày ra mà \colorbox{green}{đánh}.  

& 2                                    & 1                                 & 0                                      & 0                                    & 0                                    & 2                                
\\ (\textbf{English}: go home and fight with your son because you deserve the devil.)
\\ \hline
chả \colorbox{green}{mong} gì nhiều chỉ \colorbox{green}{mong} về \colorbox{green}{già} \colorbox{green}{được như} hai ông bà! & 0                                    & 0                                 & 2                                      & 0                                    & 1                                    & 1                                            
\\ (\textbf{English}: expecting nothing more than to be like my grandparents!)
\\ \hline
làm công nhân đã bị \colorbox{green}{vắt kiệt} \colorbox{green}{sức khỏe} cho đến khi \colorbox{green}{bị thải}. & 1                                    & 2                                 & 0                                      & 2       & 1 & 2                         
\\ (\textbf{English}: draining workers' health till they are cut loose.)
\\ \hline
\end{tabular}
\label{tab:5}
\end{table*}

Figure \ref{fig:2} illustrates how we combine sentiment analysis datasets with an emotion lexicon. The VnEmoLex is used to represent raw texts from three datasets: UIT-VSMEC, UIT-VSFC, and ViHSD to vectors as illustrated in Table \ref{tab:5}. Then, those vector are combined with the feature vectors provided by classification models for text classification such as BERT \cite{devlin-etal-2019-bert} and fastText \cite{grave-etal-2018-learning}. After that, those vectors are fit to the classification models.

\subsection{Classification models}
We present our approach for constructing the classification models in this section. In addition, according to \cite{BIRJALI2021107134}, to represent a text in vector, word embedding is a standard choice. Hence, we choose the fastText\footnote{\url{https://fasttext.cc/docs/en/crawl-vectors.html}} word embedding  which was created on Vietnamese text data \cite{grave-etal-2018-learning} as the pre-trained word embedding because this pre-trained word embedding gives good results with the social media data \cite{huynh-etal-2020-simple}. The brief details of classification models are described below.

\textbf{Text-CNN}\cite{kim-2014-convolutional}: Text-CNN is an adaption of the convolutional model used for text classification. In the Text-CNN, text data can be viewed as sequential data such as time series data or a one-dimensional matrix. Then, the convolutional layers extract important features from the texts.

\textbf{Bi-GRU}\cite{bahdanau2014neural}: GRU is a more advanced kind of Bi-LSTM, often referred to as update gate and reset gate, which was utilized to alleviate the gradient loss issue that traditional RNNs encountered. The two vectors essentially determine what data should be sent to the output. The unique feature is that it can be taught to retain old data without erasing output prediction-related data.

\textbf{BERT}\cite{devlin-etal-2019-bert}: BERT is a contextualized word representation model pre-trained
using bidirectional transformers and based on a masked language model. BERT showed power in many NLP tasks including the sentiment analysis task and is currently the SOTA method. BERT and its variants are called the BERTology \cite{rogers-etal-2020-primer}.

\textbf{PhoBERT}\cite{phobert}: PhoBERT is a monolingual language model trained on Vietnamese text. The architecture of PhoBERT is based on the RoBERTa model and is optimized for training on large-scale Vietnamese corpus. For the downstream task, the PhoBERT model requires the VnCoreNLP \cite{vu-etal-2018-vncorenlp} for pre-processing texts.  

\textbf{XLM-R}\cite{conneau-etal-2020-unsupervised}: XLM-R is a multilingual model that was developed with more than two terabytes of filtered and cleaned CommonCrawl data. Important contributions of XLM-R include up-sampling low-resource languages during training and vocabulary generation, creating a larger common vocabulary, and increasing the total model capacity to 550 million parameters.

\section{Experiments and results}
\label{ex}
\subsection{Model settings}
\textbf{Text-CNN}: We set up four conv2D layers with 32 filters at sizes 1, 2, 3, 5 and used softmax for activation. In addition, we set batch size equal to 32,
max sequence length is 100, and dropout is 0.2 for this model.

\textbf{Bi-GRU}: We set up the bidirectional layer followed by a max-pooling 1D, a dense layer has 50 in size for both activation and the softmax activation. We set batch size equal to 32, max sequence length is 80, and dropout is 0.2.

\textbf{BERTology models}: We use the multilingual BERT model (bert-base-multilingual-cased), PhoBERT-base, and XLM-R-large as pre-trained models for this approach. We set the max sequence length is 100, batch size equals 16, the learning rate is at 1e-5, accumulation steps are 5, and the number of train epochs is 4.

\subsection{Experimental Results}

\begin{table}[H]
    \caption{Experiment results on the VSMEC dataset}
    \label{tab_vsmec_no_pp}
    \resizebox{.5\textwidth}{!}{
    \begin{tabular}{p{2cm}ccc}
        \hline
        \textbf{Models} & \textbf{Accuracy (\%)} & \textbf{Macro F1-score (\%)} & \textbf{Weighted F1-score (\%)} \\ \hline
        Text-CNN & 54.83 & 52.53 & 54.77 \\
        Bi-LSTM & 52.57 & 51.22 & 52.60 \\
        Bi-GRU & 52.83 & 52.03 & 53.01\\
        BERT & 58.58 & 55.66 & 58.53 \\ 
        PhoBERT & 62.77  & 61.84 & 62.92 \\
        \textbf{XLM-R} & \textbf{67.53} & \textbf{62.79} & \textbf{67.22} \\
        \hline
    \end{tabular}
    }
\end{table}

We present our experimental results based on the procedure as mentioned in Section \ref{method}. First of all, we evaluate the impact of the pre-processing technique on the performance of the model. According to the experimental results in Table \ref{tab_vsmec_no_pp}, the XLM-R gives the best results with 62.79\% by macro F1 score on the UIT-VSMEC without the VnEmoLex. Therefore, we use the XLM-R model as the standard model to evaluate the effectiveness of pre-processing techniques. 

The results of the techniques used to experiment with the XLM-R model are summarized in Table \ref{tab:8}. From the results, the three methods (1+2+3) give the best results on the UIT-VSMEC dataset, which is 64.80\% by macro F1-score. Hence, we apply these three methods to pre-process the text before fitting it to the classification models in combination with the VnEmoLex. 

\begin{table}[H]
\centering
\caption{Data pre-processing techniques for XLM-R and their impact on the UIT-VSMEC dataset}
\resizebox{.5\textwidth}{!}{
    \begin{tabular}{lccc}
    \hline
    \textbf{Technique} & \multicolumn{1}{l}{\textbf{Accuracy (\%)}} & \multicolumn{1}{l}{\textbf{Macro F1-Score (\%)}} & \multicolumn{1}{l}{\textbf{Weighted F1-Score (\%)}} \\ \hline
    Original & 67.53 & 62.79 & 67.22 \\
    1 & 67.68 & 62.84 & 67.41 \\
    1+2 & 67.97 & 63.07 & 67.48 \\
    1+2+3 & \textbf{68.83} & \textbf{64.80} & \textbf{68.50} \\
    1+2+3+4 & 68.53 & 64.22 & 68.13 \\
    1+2+3+4+5 & 67.92 & 63.50 & 67.38 \\
    2+3 & 68.09 & 64.12 & 67.81 \\
    1+3+4 & 68.78 & 64.49 & 68.20 \\
    1+3+5 & 68.60 & 64.17 & 67.93 \\ \hline
    \end{tabular}
    }
    \label{tab:8}
\end{table}

After applying three pre-processing methods, we integrate the VnEmoLex with the classification models as described in Section \ref{method}. Table \ref{tab:7} describes the results of classification models on the VSMEC dataset. It can be seen that the Text-CNN model when applying the VnEmoLex achieves the best results for deep learning models, which is 55.25\% by macro F1-score. Also. XLM-R with the VnEmolex obtains the best results with 67.03\% by macro F1-score (this result is higher than the baseline model in \cite{vsmec}). Both methodologies increase the results when applying the VnEmoLex, proving the effectiveness of the method when used with the VnEmolex data set.

\begin{table}[H]
\centering
    \caption{Evaluation results on UIT-VSMEC dataset.}
    \resizebox{.5\textwidth}{!}{
    \begin{tabular}{lccc}
        \hline
        \textbf{Models} & \textbf{Accuracy(\%)} & \textbf{Macro F1-Score(\%)} & \textbf{Weighted F1-Score(\%)} \\ 
        \hline
        Baseline model \cite{vsmec} & 59.74 & - & 59.74 \\
        \hline
        Text-CNN  & 53.79 & 53.79 & 55.95 \\
        \textbf{Text-CNN+VnEmoLex}  & \textbf{58.68} & \textbf{55.27} & \textbf{58.47} \\ 
        \hline
        Bi-LSTM  & 53.54 & 51.68 & 53.28 \\
        Bi-LSTM+VnEmoLex  & 54.40 & 53.80 & 55.28 \\ 
        \hline
        Bi-GRU  & 53.70 & 52.32 & 53.54 \\
        Bi-GRU+VnEmoLex  & 55.08 & 54.27 & 54.67 \\ 
        \hline
        BERT  & 60.12 & 57.02 & 60.25 \\
        BERT+VnEmoLex  & 60.43 & 57.80 & 60.60 \\ 
        \hline
        PhoBERT  & 64.65 & 62.32 & 65.84 \\
        PhoBERT+VnEmoLex  & 67.02 & 63.70 & 66.76 \\ 
        \hline
        XLM-R  & 68.83 & 64.80 & 68.50 \\
        \textbf{XLM-R+VnEmoLex}  & \textbf{70.42} & \textbf{67.03} & \textbf{70.06} \\ 
        \hline
    \end{tabular}
    }
\label{tab:7}
\end{table}

\begin{table}[H]
\centering
    \caption{Evaluate the results from the UIT-VSFC and ViHSD datasets.}
    \resizebox{.5\textwidth}{!}{
    \begin{tabular}{lccc}
        \hline
        \textbf{Models} & \textbf{Accuracy(\%)} & \textbf{Macro F1-Score(\%)} & \textbf{Weighted F1-Score(\%)} \\ 
        \hline
        \multicolumn{4}{c}{\textbf{UIT-VSFC}} \\
        \hline
        Baseline model \cite{vsfc} & - & - & 87.94 \\
        \hline
        Text-CNN  & 87.40  & 67.31  & 86.11  \\
        \textbf{Text-CNN+VnEmoLex}  & \textbf{89.12} & \textbf{69.74} & \textbf{88.37} \\ 
        \hline
        XLM-R  & 94.13 & 83.09  & 93.74 \\
        \textbf{XLM-R+VnEmoLex}  & \textbf{94.25} & \textbf{83.40} & \textbf{93.97} \\ 
        \hline
        \multicolumn{4}{c}{\textbf{ViHSD}} \\
        \hline
        Baseline model \cite{vihsd} & 86.88 & - & 62.69 \\
        \hline
        Text-CNN  & 85.34  & 59.66   & 85.48  \\
        \textbf{Text-CNN+VnEmoLex}  & \textbf{87.14} & \textbf{63.06} & \textbf{86.94} \\ 
        \hline
        XLM-R  & 88.05  & 66.84  & 87.75 \\
        \textbf{XLM-R+VnEmoLex}  & \textbf{88.29 } & \textbf{68.29} & \textbf{88.08} \\ 
        \hline
    \end{tabular}
    }
\label{tab:9}
\end{table}

Beside the UIT-VSMEC, we conduct the experiment on the two datasets including UIT-VSFC and ViHSD. We use the XLM-R and Text-CNN with the three pre-processing techniques (1+2+3) on both datasets. Table \ref{tab:9} summarizes all of the results. It can be seen that the XLM-R model in combining with the VnEmoLex also shows the highest results on the UIT-VSFC and ViHSD datasets, which are 83.40\% and 68.29\% by macro F1-score, respectively. 


\subsection{Error analysis}

\begin{figure}[H]
    \centering
    \includegraphics[scale=0.28]{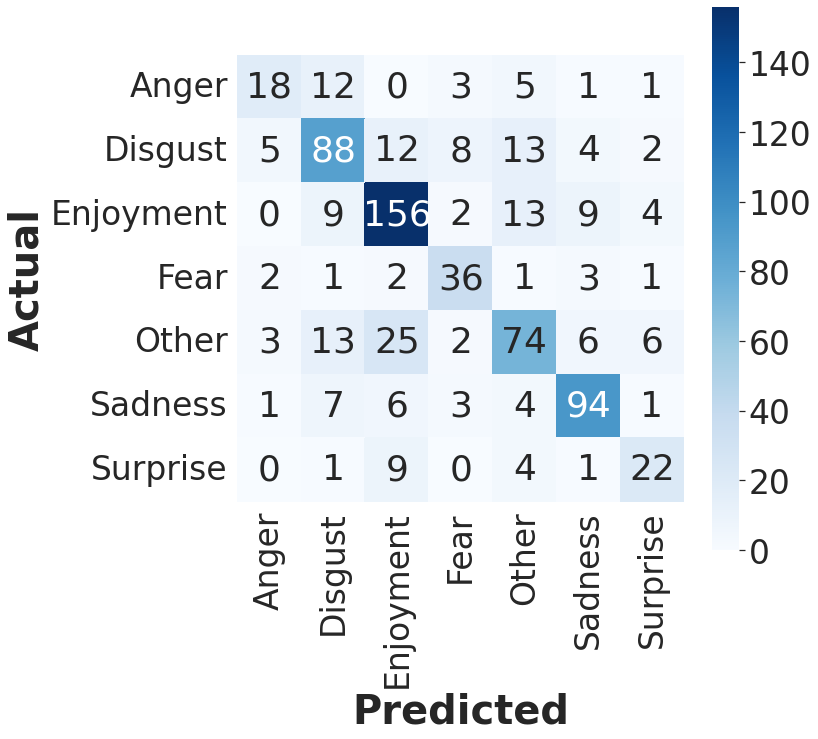}
    \caption{Confusion matrix of XLM-R on the VSMEC dataset.}
    \label{fig:4}
\end{figure}

\begin{table*}[htbp]
    \centering
    \caption{SOME ERROR CASES}
    \label{tab:10}
    \resizebox{.95\textwidth}{!}{
    \begin{tabular}{llll}
        \hline
        \textbf{Comments} & \textbf{Emotion} & \textbf{Predictions} & \textbf{Explanations} \\ \hline
        \begin{tabular}[c]{@{}l@{}}chia se cho ai thích thể hiện nè\\ (\textbf{English}: share with those who enjoy showing)\end{tabular} & Other & Enjoyment & \begin{tabular}[c]{@{}l@{}}The use of the word "thích" (like) confuses the meaning of the \\ sentence and gives the impression of enjoyment.\end{tabular} \\ \hline
        \begin{tabular}[c]{@{}l@{}}tao đâu biết nó kinh khủng thế này :((( \\ (\textbf{English}: i had no idea it was so terrible :(()\end{tabular} & Fear & Sadness & \begin{tabular}[c]{@{}l@{}}The ":((("emoticon have a stronger influence than the word \\ "kinh khủng" (terrible)\end{tabular} \\ \hline
        \begin{tabular}[c]{@{}l@{}}đây gọi là nghiệp vật\\ (\textbf{English}: this is known as karma)\end{tabular} & Enjoyment & Other & \begin{tabular}[c]{@{}l@{}}The user sometimes has difficulty  identifying the enjoyable nuance \\ of the sentence.\end{tabular} \\ \hline
        \begin{tabular}[c]{@{}l@{}}người ta có bạn bè nhìn vui ghê\\ (\textbf{English}:  they look so happily having friends)\end{tabular} & Sadness & Enjoyment & \begin{tabular}[c]{@{}l@{}}Our model cannot understand the context of this comment.\end{tabular} \\ \hline
        \begin{tabular}[c]{@{}l@{}}tao không sợ đi làm mệt mỏi tao chỉ sợ không có \\ niềm vui với nó\\ (\textbf{English}: i'm worried of not having fun with it,\\ not of going to work tired)\end{tabular} & Fear & Sadness & \begin{tabular}[c]{@{}l@{}}Both fear and sadness are present in this sentence, however \\ because both "sợ" (fear) and "không sợ" (not afraid) are present, \\our model chooses sadness\end{tabular} \\ \hline
        \begin{tabular}[c]{@{}l@{}}tao đéo ngờ tới trường hợp này :))\\ (\textbf{English}: I f*cking cannot know it)\end{tabular} & Surprise & Enjoyment & \begin{tabular}[c]{@{}l@{}}The term "đéo ngờ tới" (did not expect) has not appeared in the \\ train data so it cannot be recognized to expresssurpriseness.\end{tabular} \\ \hline
        \end{tabular}
    }
\end{table*}

We used the confusion matrix to evaluate how well the models performed by better visualizing the prediction and actuality of the labels. The confusion matrix of the XLM-R - the best classification model on the test set of the UIT-VSMEC is shown in Figure \ref{fig:4}. As can be seen, the best classification model on the Sadness label has more than 81\% correct prediction points, followed by the Enjoyment label's accuracy of 80.80\% and the Fear label's accuracy of more than 78\%. Label Anger has the lowest accuracy at 45\%.

We can obtain both good and bad prediction labels using the confusion matrix. For example, when predicting the relationship between the labels of Anger and Disgust, which have conceptual levels that are quite comparable, annotators frequently struggle with meaning confusion. Therefore, model confusion is unavoidable. Table \ref{tab:10} contains a few illustrations of prediction errors along with their corresponding explanations.

\section{Conclusion}
\label{conclusion}
In this paper, we propose a methodology that combines classification models with the VnEmoLex lexicons for the sentiment analysis task in Vietnamese. The results showed that the VnEmoLex lexicon has significantly improved the performance of classification models on the three datasets including the UIT-VSMEC, UIT-VSFC, and ViHSD. The XLM-R combined with the VnEmoLex lexicons obtains the highest results on three datasets, which are 67.03\% for the UIT-VSMEC, 83.40\% for the UIT-VSFC, and 68.29\% for the ViHSD by macro F1-score. Besides, the text pre-processing techniques also play an important role in boosting the classification model. In addition, the ambiguity in the text caused by the use of social languages such as abbreviation, metaphor, and emoticon makes the classification models confused in detecting the actual emotion from the text. In the future, we will improve the model performance by using the semantic labels and grammatical rules for the sentiment task to solve the problem of ambiguity in social media texts.

\bibliographystyle{IEEEtran}
\bibliography{references}

\end{document}